\documentclass[letterpaper, 10 pt, conference]{ieeeconf}  

\IEEEoverridecommandlockouts                              

\usepackage{graphicx}
\usepackage{hyperref}
\usepackage{multirow}
\usepackage{tabularx}
\usepackage{booktabs,subcaption,amsfonts,dcolumn}
\usepackage{listings}
\usepackage{color}
\definecolor{backcolour}{rgb}{0.95, 0.95, 0.96}
\usepackage[dvipsnames]{xcolor}
\usepackage{tablefootnote}

\definecolor{codegreen}{rgb}{0,0.6,0}
\definecolor{codegray}{rgb}{0.5,0.5,0.5}
\definecolor{codepurple}{rgb}{0.58,0,0.82}
\definecolor{backcolour}{rgb}{0.95,0.95,0.92}

\newcommand{\lidaraug}{LidarAugment}

\usepackage{graphics} 
\usepackage{epsfig} 
\usepackage{mathptmx} 
\usepackage{times} 
\usepackage{amsmath} 
\usepackage{amssymb}  

\title{\LARGE \bf \lidaraug{}: Searching for Scalable 3D LiDAR Data Augmentations
}

\author{Zhaoqi Leng$^{1*}$, Guowang Li$^{1}$, Chenxi Liu$^{1}$, Ekin Dogus Cubuk$^{2}$, Pei Sun$^{1}$, Tong He$^{1}$, \\
Dragomir Anguelov$^{1}$  and Mingxing Tan$^{1}$
\thanks{$^{1}$ Waymo Research, $^{2}$ Google Brain, $^{*}$ lengzhaoqi@waymo.com}%
}

\begin{document}

\maketitle
\thispagestyle{empty}
\pagestyle{empty}

\begin{abstract}

Data augmentations are important in training high-performance 3D object detectors for point clouds. Despite recent efforts on designing new data augmentations, perhaps surprisingly, most state-of-the-art 3D detectors only use a few simple data augmentations. In particular, different from 2D image data augmentations, 3D data augmentations need to account for different representations of input data and require being customized for different models, which introduces significant overhead. In this paper, we resort to a search-based approach, and propose \emph{\lidaraug{}}, a practical and effective data augmentation strategy for 3D object detection. Unlike previous approaches where all augmentation policies are tuned in an exponentially large search space, we propose to factorize and align the search space of each data augmentation, which cuts down the 20+ hyperparameters to 2, and significantly reduces the search complexity.
We show \lidaraug{} can be customized for different model architectures with different input representations by a simple 2D grid search, and consistently improve both convolution-based UPillars/StarNet/RSN and transformer-based SWFormer. 
Furthermore, \lidaraug{} mitigates overfitting and allows us to scale up 3D detectors to much larger capacity. In particular, by combining with latest 3D detectors, our \lidaraug{} achieves a new state-of-the-art 74.8 mAPH L2 on Waymo Open Dataset.
\end{abstract}

\section{Introduction}

Data augmentations are widely used in training deep neural networks. In particular, for autonomous driving, many data augmentations are developed to improve data efficiency and model generalization. However, most recent 3D object detectors only use a few basic data augmentation operations such as rotation, flip and ground-truth sampling \cite{yan2018second,lang2019pointpillars,zhou2020end,shi2020pv,liang2020rangercnn,sun2021rsn,bewley2021range}. This is in a surprising contrast to 2D image recognition and detection, where much more sophisticated 2D data augmentations are commonly used in modern image-based models~\cite{zhang2017mixup,cubuk2018autoaugment,yun2019cutmix, ho2019population,cubuk2020randaugment,zhong2020random}. In this paper, we aim to answer: \emph{is it practical to adopt more advanced 3D data augmentations to improve modern 3D object detectors, especially for high-capacity models?}

\begin{figure}[t!]
    \centering
    \includegraphics[width=0.8\columnwidth]{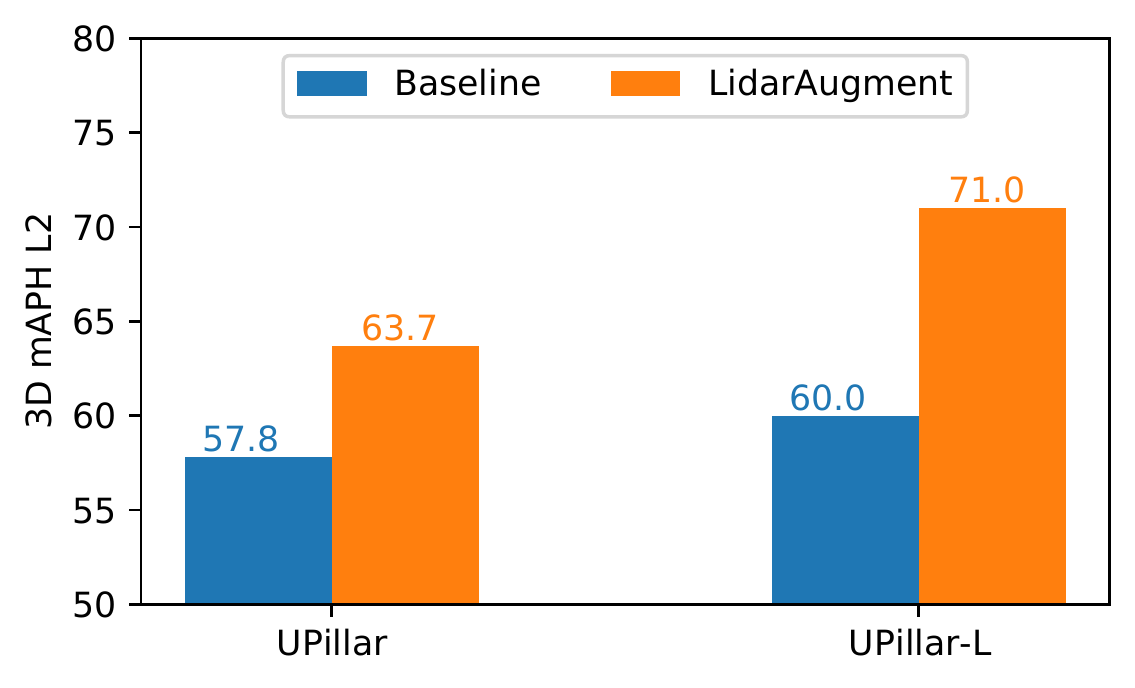}
    \vspace{-0.1in}
    \caption{\textbf{Model scaling with \lidaraug{} on Waymo Open Dataset.} Baseline augmentations are from the prior art of \cite{swformer}. When scaling up UPillars to UPillars-L, our \lidaraug{} improves both models, and the gains are more significant for the larger model, thanks to its customizable regularization.
    More results in \autoref{tab:upillar-scale}.}
    \label{fig:scaling}
    \vspace{-0.1in}
\end{figure}

\begin{figure*}[t!]
    \centering
    \vspace{2mm}
    \includegraphics[width=1.9\columnwidth]{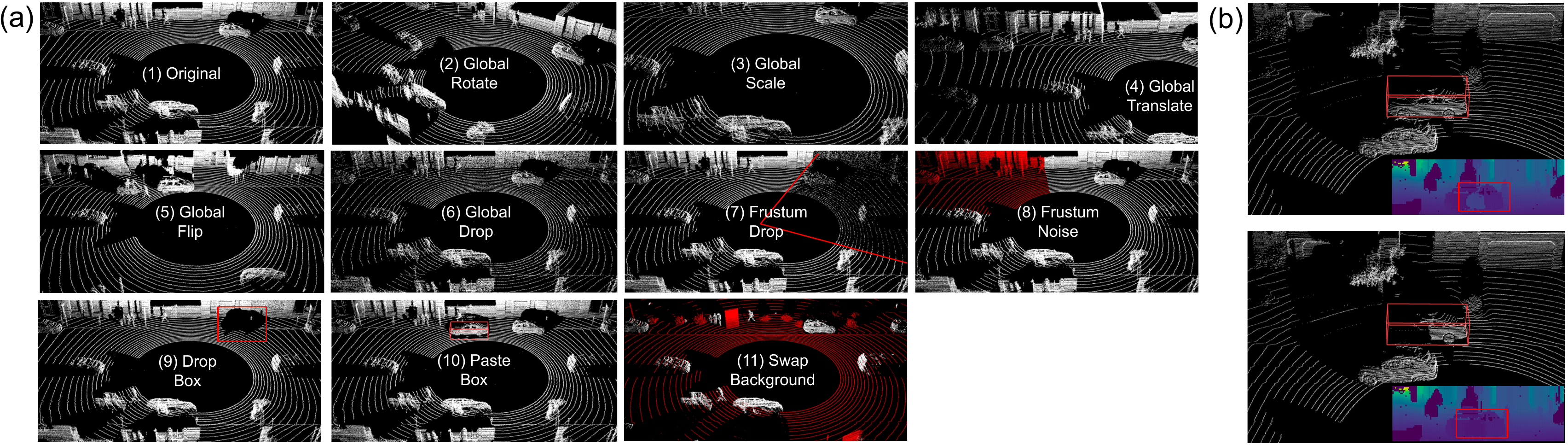}
    \caption{\textbf{Visualizing \lidaraug{}.} (a) all data augmentation operations used in \lidaraug{}. For non-global operations, we highlight the augmented parts in red (boxes). (b) occlusion introduced by data augmentation, e.g., paste a car object, is handled by removing overlapping rays in range view based on distance.
    We show point clouds and the corresponding range images with (bottom)/without (top) removing overlapping rays in the range view.
    }
    \label{fig:coherent_ops}
\end{figure*}

The main challenge of adopting advanced 3D data augmentations is that 3D augmentations are often sensitive to input representations and model capacity. For example, range image based models and point cloud based models require different types of data augmentation due to different input representations. High capacity 3D detectors are typically prone to overfitting and require stronger overall data augmentation compared to lite models with fewer parameters. Therefore, tailoring each 3D augmentation for different models is necessary. However, the search space scales exponentially with respect to the number of hyperparameters, which leads to significant search cost. Recent studies~\cite{cheng2020improving,li2020pointaugment} attempt to address these challenges by using efficient search algorithms. Those approaches typically construct a fixed search space, and run a complex search algorithms (such as population-based search \cite{jaderberg2017population}) to find a data augmentation strategy for a model. However, our studies reveal that the search spaces used in prior works are suboptimal. Despite having complex search algorithms, without a systematic way to define a good search space, we cannot unleash the potential of a model. 

In this paper, we propose \lidaraug{}, a simplified search-based approach for 3D data augmentations. Unlike previous methods that rely on complex search algorithms to explore an exponentially large search space, our approach aims to define a simplified search space that contains a variety of data augmentations but has minimal (i.e. two) hyperparameters, such that users can easily customize a diverse set of 3D data augmentations for different models. 

Specifically, we construct the \lidaraug{} search space by first factorizing a large search space based on operations and exploring each sub search space with a per-operation search. Then, we normalize and align the sub search space for each data augmentation to form the \lidaraug{} search space. The final \lidaraug{} search space contains only two shared hyperparameters: $m \in [0, \infty)$ controls the normalized magnitude and  $p \in [0, 1]$ controls the probability of applying each data augmentation policies. Our \lidaraug{} search space significantly simplifies prior works \cite{cheng2020improving} by cutting down the number of hyperparameters to two, a 15$\times$ reduction in number of hyperparameters. 

Despite only having two hyperparamters, our \lidaraug{} search space contains a variety of existing 3D data augmentations, such as drop/paste 3D bounding boxes, rotate/scale/dropping points, and copy-paste objects and backgrounds. In addition, \lidaraug{} supports coherent augmentation across both point and range view representations, which generalizes to multi-view 3D detectors.


We perform extensive experiments on the Waymo Open Dataset \cite{sun2020scalability} and demonstrate \lidaraug{} is effective and generalizes well to different model architectures (convolutions-based and transformer-based), different input views (3D point view and range image), and different temporal scales (single and multi frames). Notably, \lidaraug{} advances state-of-the-art (SOTA) transformer-based SWFormer by 1.4 mAPH on the test set. Furthermore, \lidaraug{} provides customizable regularization, which allows us to scale up 3D object detectors to much higher capacity without overfitting. As summarized in \autoref{fig:scaling}, \lidaraug{} consistently improves UPillars models, and the performance gains are particularly large for high-capacity models.  Our contributions can be summarized as:
\begin{enumerate}
    \item \textbf{New insight:} we reveal that common 3D data augmentation search spaces are suboptimal and should be tailored for different models. 
    \item \textbf{\lidaraug{}:} we propose the \lidaraug{} search space, which supports jointly optimizing 10 augmentation policies with only two hyperparameters (15$\times$ reduction compares to prior works), offering diverse yet practical augmentations. In addition, we develop a new method to coherently augment both point and range-view input representations.
    \item \textbf{State-of-the-art performance:} \lidaraug{} consistently improves both convolution-based UPillars/StarNet/RSN and attention-based SWFormer. With \lidaraug{}, we achieve new state-of-the-art results on Waymo Open Dataset. In addition, \lidaraug{} enables model scaling to achieve much better quality for high-capacity 3D detectors.
\end{enumerate}

\section{Related works}
\textbf{Data augmentation.} Data augmentation is widely used in training deep neural networks 
In particular, for 3D object detection from point clouds, several global and local data augmentations, such as rotation, flip, pasting objects, and frustum noise, are used to improve model performance ~\cite{chen2017multi,yan2018second,yang2018pixor,lang2019pointpillars,shi2020pv,chen2020pointmixup,cheng2020improving,hu2021pattern,choi2021part,reuse2021ambiguity}. However, as 3D data augmentations are sensitive to model architectures and capacity, it often requires extensive manual tuning to use these augmentations. Therefore, most existing 3D object detectors \cite{lang2019pointpillars, sun2021rsn, yin2021centerpoint, singlestride21, swformer} only adopt a few simple augmentations, such as flip and shift pixels.

Several recent works attempt to use range images for multi-view 3D detection, but very few augmentations are developed for range images. \cite{liang2020rangercnn} attempts to paste objects in the range image without handling occlusions. Our Paste Box augmentation support coherently augmenting both range-view and point-view input data while handling occluded objects in a simple way (more details in \autoref{fig:coherent_ops}), which enables more realistic augmented scenes and enriches the data augmentations for multi-view 3D detectors.

\textbf{Learning data augmentation policies.} 
Designing good data augmentation normally requires manual tuning and domain expertise. Several search-based approaches have been proposed for 2D images, such as AutoAugment \cite{cubuk2018autoaugment},  RandAugment \cite{cubuk2020randaugment}, and Fast AutoAugment~\cite{lim2019fast}. Our \lidaraug{} is inspired by RandAugment in the sense that we both try to construct a simplified search space. However, unlike 2D image augmentations, where a search space works well for many models, we reveal that existing search space for 3D detection tasks are suboptimal, which motivates us to propose the first systematical method to define search spaces for 3D detection tasks.

On the other hand, for 3D detection, PPBA \cite{cheng2020improving} and PointAugment \cite{li2020pointaugment} propose efficient learning-based data augmentation frameworks for 3D point clouds. However, both works require users to run a complex algorithm on an exponentially large but not well-designed search space. In contrast, our work provides a systematical framework to design a simple and more effective search spaces with only two hyperparameters. 

\section{\lidaraug{}}
In this section, we first introduce data augmentation policies used in \lidaraug{}. Next, we analyze the performance of each data augmentation policy on Waymo Open Dataset \cite{sun2020scalability}. Finally, we propose a systematic approach to progressively design 3D augmentation search space. 

\subsection{Data augmentations for point clouds and range images.} \label{subsec:range}
3D point cloud and 2D range image are two different representations of LiDAR data. Despite being the native representation of LiDAR data, data augmentations for range image is not well studied compared to point clouds. Here, we revisit data augmentations for point clouds, and introduce a new method for coherently applying data augmentation to both point clouds and range images.

\textbf{Augmenting point clouds.} We follow the implementation of data augmentation policies described in recent studies \cite{yan2018second,cheng2020improving,leng2022pseudoaugment}, which contain global operations (rotate, scale, translate, flip, and drop points) and local operations (drop boxes, paste boxes, swap background, drop points and add feature noise in a frustum), shown in \autoref{fig:coherent_ops} (a).

\textbf{Augmenting range images.} Different from sparse 3D point representation, pixels in range image are compact. Data augmentations, such as pasting objects and swap background, disturb the compact structure of range representation. Here, we propose a novel approach to coherently augment both 3D point view and 2D range view by leveraging the bijective property between point clouds and range images, while account for occlusion. 

First, we transform the range image pixels to point cloud based on $(x, y, z)$ coordinates. To preserve the bijective mapping between a pixel in a range image and a point in the corresponding point clouds, we concatenate the (row, column) index of each pixel in the range image as additional features before scattering pixels to 3D. After performing data augmentation in the point representation, we transform the augmented point clouds back to the range view by scattering each point to a pixel in a 2D image based on its (row, column) index.

\textbf{Leveraging the compactness of range images.} Coherently augmenting both range and point views leads to more realistic augmented scenes. Because each pixel in a range image corresponds to a unique ray from LiDAR, overlapping pixels in the range view represent that the same light ray penetrates trough multiple surfaces. When this happens, we compare the distance among overlapping pixels in the range view and keep the pixel that is closest to the ego vehicle. This effectively removes occluded points in both the range and point views, as shown in \autoref{fig:coherent_ops} (b). 

\begin{table}
\vspace{2mm}
\resizebox{0.98\columnwidth}{!}{
\begin{tabular}{l p{3.1cm} |c|l}
\toprule
Policy & Hyperparameters & WOD (Veh./Ped.) & mAP L1 \\
\midrule
No Aug & -& - & 60.2\\
\midrule
\multirow{2}{*}{Drop Box} & Probability  & $p/p$ & \multirow{2}{*}{66.0 \small\textcolor{blue}{(+5.8)}}  \\
& Number of boxes & $2 m/2.8 m$ & \\
\midrule
\multirow{3}{*}{Paste Box} & Probability  & $1.4 p/p$ & \multirow{2}{*}{66.6 \small\textcolor{blue}{(+6.4)}}\\
& Number of boxes & $3.2 m/4.4  m$ & \\
\midrule
Swap Background & Probability  & $0.6  p $ & 63.6 \small\textcolor{blue}{(+3.4)}\\
\midrule
\multirow{2}{*}{Global Rot} & Probability  & $1.4  p$ & \multirow{2}{*}{73.3 \small\textcolor{blue}{(+13.1)}}\\
& Max rotation angle & $0.22\pi  m $ & \\
\midrule
\multirow{2}{*}{Global Scale} & Probability  & $p$ & \multirow{2}{*}{66.0 \small\textcolor{blue}{(+5.8)}}\\
& Scaling factor & $0.036  m$ & \\
\midrule
\multirow{2}{*}{Global Drop} & Probability  & $p$ &  \multirow{2}{*}{64.9 \small\textcolor{blue}{(+4.7)}}\\
& Drop ratio & $1 - 0.18  m$ & \\
\midrule
\multirow{5}{*}{Frustum Drop} & Probability  & $p$ & \multirow{5}{*}{64.1 \small\textcolor{blue}{(+3.9)}}\\
& Theta angle width & $0.1 \pi  m$ & \\
& Phi angle width &  $0.1 \pi  m$  & \\
& R distance & $75 - 7.5m$ & \\
& Drop ratio & $1-0.1m$ & \\
\midrule
\multirow{5}{*}{Frustum Noise} & Probability  & $0.6 p$ & \multirow{5}{*}{65.1 \small\textcolor{blue}{(+4.9)}} \\
& Theta angle width & $0.14 \pi m$ & \\
& Phi angle width & $0.14 \pi m$ & \\
& R distance & $75- 10.5 m$ & \\
& Max noise level & $0.14 m$ & \\
\midrule
\multirow{2}{*}{Global Translate} & Probability & $1.4p$ & \multirow{2}{*}{67.5 \small\textcolor{blue}{(+7.3)}}\\
& Stdev. of noise (x, y)&  $0.66 m$ & \\
\midrule
Global Flip &  Probability & $p$ & 69.0 \small\textcolor{blue}{(+8.8)}\\
\bottomrule
\end{tabular}}
\caption{\textbf{Aligned search spaces and performance.} The search space of each hyperparameter for Waymo Open Dataset (WOD) for UPillars is listed. $(p, m)$ are two global hyperparameters to control all data augmentation policies. After aliging the search space, the optimal $(p, m)$ for each data augmentation are $(0.5, 5)$. The probability of each policy is clipped to [0, 1]. The min R distance is clipped to 0. The maximum rotation angle is clipped to [0, $\pi$]. The maximum flip probability is clipped to 0.5. The ratio of dropped points are clipped to [0, 0.8]. The theta angle and phi angle are clipped to [0, $\pi$] and [0, $2\pi$], respectively.  }
\label{tab:searchspace}
\vspace{-0.1in}
\end{table}

\begin{figure*}[h!]
    \centering
    \vspace{2mm}

    \includegraphics[width=0.95\textwidth]{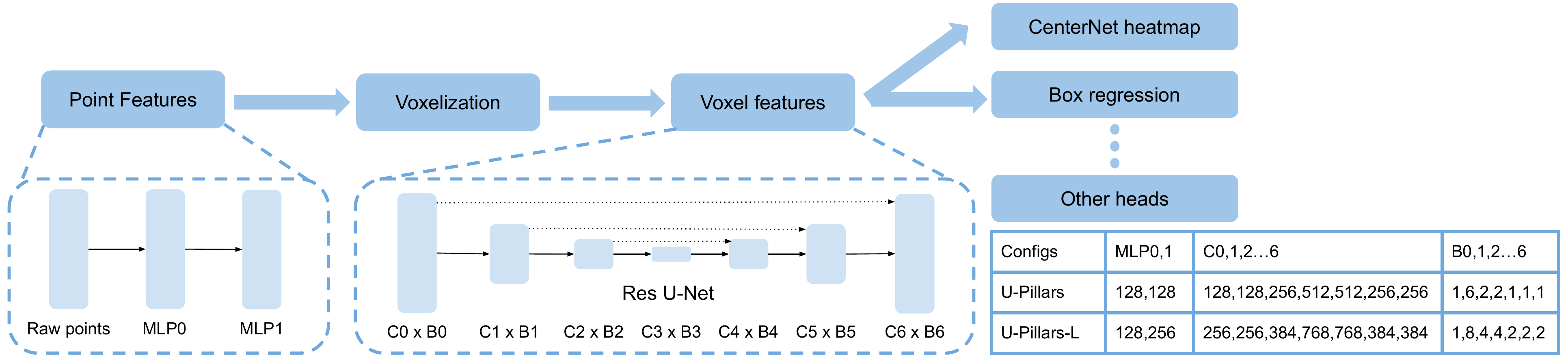}
    \caption{\textbf{UPillars architecture.} Input points are processed by two full connected layers with channel size (MLP0, MLP1) before voxelized into pillars. The bird-eye-view pillars are processed by a Res U-Net, where the channel size and number of blocks at each resolution are (Ci, Bi). CenterNet detection head, box regression, and other attributes regression heads are applied to the output of U-Net. For both models, we use the same voxel size $0.32 m$ and range $81.92 m$.}
    \label{fig:UPillars}
\end{figure*}

\subsection{Effects of each data augmentation.} \label{sec:effects_augmentation}
In this section, we assess the effects of each data augmentation policy on Waymo Open Dataset \cite{sun2020scalability}. To benchmark the policies, we develop a UPillars architecture, which is based on the popular PointPillars \cite{lang2019pointpillars}, but incorporates recent optimizations in architecture design, i.e., unet backbone \cite{ronneberger2015u}, and center net detection head \cite{yin2021center}.

\textbf{Datasets and training.} Waymo Open Dataset \cite{sun2020scalability} contains 798 and 202 training and validation sequences. For the following studies, we train UPillars with batch size 64, Adam optimizer \cite{kingma2014adam} and cosine decay learning rate with max learning rate 3e-3 and total step 80000.

\textbf{Effect of each data augmentations.} We factorize the \lidaraug{} search space into per-policy sub search space and show the UPillars performance when trained using only one policy on Waymo Open Dataset (WOD) in \autoref{tab:searchspace}. Interestingly, on WOD, the most effective data augmentation technique is global rotation, whereas on KITTI \cite{geiger2013vision}, pasting ground truth bounding boxes is commonly regarded as the most effective data augmentation \cite{yan2018second}. A closer look at the statistics of the two datasets reveals that, on average, each KITTI LiDAR frame contains about five objects, whereas, each frame in WOD on average contains more than 50 objects. Thus, pasting ground truth objects has larger impact on KITTI, due to the significantly lower object density, than WOD. On the other hand, smaller global rotation angle $\pi/4$ is commonly used when training KITTI dataset, but we find much stronger rotation $\pi$ is preferred for WOD.  

\subsection{Defining \lidaraug{} search space.} \label{sec:define_searchspace}
As indicated from the previous section, different from RandAugment for 2D images \cite{cubuk2020randaugment}, naively using the same search space across different datasets is suboptimal, which is a unique challenge for 3D detection tasks. To mitigate this new challenge, we propose to factorize the whole search space and align each data augmentation based on its optimal hyperparameters.

\noindent \textbf{Align the search space.} Using global hyperparameters to control all data augmentations requires normalizing the search domain of each hyperparameter. Without normalization, the same global magnitude could lead to an aggressive application of one data augmentation, and an insufficient application of another data augmentation. To align the search domain of each data augmentation policy, we train a UPillars model on a given dataset while only applying a single data augmentation policy at each time. Based on the optimal values of hyperparameters, we rescale the search domain of each hyperparameter in a data augmentation policy from  $[0, arbitrary\_value]$ to $[0, optimal\_value]$ such that the optimal value for each hyperparameter corresponds to the same global magnitude or probability hyperparameter. Since each data augmentation policy contains multiple parameters, to save cost, we perform a small scale 2D grid search to scale the probability and magnitudes of all hyperparameters in each sub search space.

\lstdefinestyle{mystyle}{
    backgroundcolor=\color{backcolour},   
    commentstyle=\color{codegreen},
    keywordstyle=\color{magenta},
    numberstyle=\tiny\color{codegray},
    stringstyle=\color{codepurple},
    basicstyle=\ttfamily\footnotesize,
    breakatwhitespace=false,         
    breaklines=true,                 
    captionpos=b,                    
    keepspaces=true,                 
    numbers=left,                    
    numbersep=5pt,                  
    showspaces=false,                
    showstringspaces=false,
    showtabs=false,                  
    tabsize=2
}
\lstset{style=mystyle}

\begin{figure}[!b]
\vspace{-0.2in}
\begin{minipage}{\columnwidth}
\begin{lstlisting}[language=Python,numbers=none]
augmentations = [
    DropBox, PasteBox, SwapBackground, GlobalRot, 
    GlobalScale, GlobalDrop, FrustumDrop ,
    FrustumNoise, GlobalTranslate, GlobalFlip]
def lidaraugment(m, p, input_frame):
    for aug in augmentations:
        aug.set_magnitude(m)
        aug.set_probability(p)
        input_frame = aug.transform(input_frame)
    return input_frame
\end{lstlisting}
\end{minipage}
\vspace{-0.1in}
\caption{\textbf{Pseudo Python code for \lidaraug{}.}}
\label{fig:pseudocode}
\end{figure}

Here, we use Global Translate as an example. We define the initial search domain for the probability of applying Global Translate data augmentation to be $\{0.3, 0.5, 0.7, 0.9\}$, and the domain for the magnitude of translation noise to be $\{0.9, 1.5, 2.1, 2.7, 3.3, 3.9\}$. If the optimal values $(p_{\text{noise}}, m_{\text{noise}}) = (0.7, 3.3)$, we rescale the search domain to $(p_{\text{noise}}, m_{\text{noise}}) = (1.4p, 0.66m)$ such that when the global hyperparameters $(p, m) = (0.5, 5)$, the hyperparameters for Global Translate are optimal. Details about all the hyperparameters are listed in \autoref{tab:searchspace}. The \lidaraug{} pseudocode is shown in \autoref{fig:pseudocode}.
\section{Experiments}

In this section, we first introduce our experimental setups. Then we show \lidaraug{} significantly improves the performance for both convolution-based and attention-based models. Lastly, we show the model scaling results, followed by ablations studied on different models and datasets.

\subsection{Experimental setup}
Our experiments are mostly based on Waymo Open Dataset~\cite{sun2020scalability} (WOD) where the main metric is mAPH L2, with additional ablation studies on nuScene~\cite{caesar2020nuscenes}. We evaluate \lidaraug{} on a variety of 3D object detectors, as well as different model sizes. For fair comparison, we strictly follow the original training settings for each model, and only replace the baseline augmentation with our new \lidaraug{}. We train UPillars using Adam optimizer \cite{kingma2014adam} and apply cosine learning rate with max learning rate 1e-3, total 16e4 steps and batch size 64. 

\begin{table*}[!t]
\vspace{2mm}
\centering
 \resizebox{0.6\textwidth}{!}{
\begin{tabular}{lc|c|cc|ccc}
\toprule
\multirow{2}{*}{Method} &
\multirow{2}{*}{Type} &
\multirow{1}{*}{mAPH} &
\multicolumn{2}{c|}{Vehicle AP/APH 3D} &
\multicolumn{2}{c}{Pedestrian AP/APH 3D} \\
& & L2 & L1 & L2 & L1 & L2 \\
\midrule
\midrule
P.Pillars~\cite{lang2019pointpillars} \dag & conv & 51.9 & 63.3/62.7 &  55.2/54.7 &  68.9/56.6 & 60.4/49.1 \\
CenterPoint~\cite{yin2021centerpoint}  & conv & 67.1 & 76.6/76.1 & 68.9/68.4 & 79.0/73.4 & 71.0/65.8 \\
RSN\_3f~\cite{sun2021rsn}  &conv & 68.1 & 78.4/78.1 & 69.5/69.1 & 79.4/76.2 & 69.9/67.0 \\
PVRCNN++~\cite{shi2021pv++}  &conv & 69.1 & 79.3/78.8 & 70.6/70.2 & 81.8/76.3 & 73.2/68.0  \\
UPillars-L$^\dag$ &conv &  60.0& 69.5/69.0 & 61.5/61.0 & 70.4/66.1 &  63.0/59.0\\
\textbf{UPillars-L(+LA)}   &\textbf{conv} &  \textbf{71.0}& \textbf{79.5/79.0} & \textbf{71.9/71.5} & \textbf{81.5/77.3} & \textbf{74.5/70.5} \\
\midrule
\midrule
SST\_1f \cite{singlestride21} & attn & 63.4 &74.2/73.8 & 65.5/65.1 & 78.7/69.6 & 70.0/61.7 \\
SST\_3f \cite{singlestride21} & attn & 69.5 &77.0/76.6 & 68.5/68.1 & 82.4/78.0 & 75.1/70.9 \\
SWFormer \cite{swformer}  & attn  & 70.9 & 79.4/78.9 & 71.1/70.6 & 82.9/79.0 & 74.8/71.1 \\
\textbf{SWFormer(+LA)}  & \textbf{attn} &\textbf{72.8}& \textbf{80.9/80.4} & \textbf{72.8/72.4}& \textbf{84.4/80.7} &\textbf{76.8/73.2} \\

\bottomrule
\end{tabular}
}
\centering
\caption{\textbf{WOD validation-set results.} \emph{LA} denotes our \lidaraug{}, \emph{conv} denotes convolutional networks, and \emph{attn} denotes attention-based transformer models. \lidaraug{} improves both types of models, and achieves the best results among each category. $^\dag$ model is trained using augmentations shown in prior art \cite{swformer}.}
\label{tab:validation}
\end{table*}

\subsection{\lidaraug{} achieves new state-of-the-art results} \label{sec:sota_swf}

\autoref{tab:validation} compares the validation set results on Waymo Open dataset. Our \lidaraug{} significantly improves both convolution-based and transformer-based models. In particular, by scaling up the basic UPillars, our \lidaraug{} achieves 71.0 mAPH of L2 on UPillars-L, which is 1.9 AP better than the previous best convolution-based 3D detector PVRCNN++ \cite{shi2021pv++}. Notably, the latest transformer-based SWFormer~\cite{swformer} already uses 4 strong data augmentation policies, i.e., rotation (probability 0.74, yam angle uniformly sampled from [$-\pi$,  $\pi$]), random flip (probability 0.5), randomly scaling the world (scaling factor uniformly sampled from $[0.95, 1.05]$, and randomly drop points (drop probability 0.05), where the rotation angle and flip probability are maxed out. Despite that, \lidaraug{} still outperforms SWFormer by 1.9 AP, establishing a new state-of-the-art result for single-modal models without ensemble or test time augmentation on Waymo Open Dataset.

\autoref{tab:test_result} compares the test-set results among latest models. Compared to the latest SWFormer, our \lidaraug{} improves the test-set L2 mAPH by 1.4 AP, outperforming all prior arts by a large margin.

\begin{table}[!h]
\centering
 \resizebox{0.5\textwidth}{!}{
\begin{tabular}{l|c|cc|ccc}
\toprule
\multirow{2}{*}{Method} &
\multirow{1}{*}{mAPH} &
\multicolumn{2}{c}{Vehicle AP/APH 3D} &
\multicolumn{2}{c}{Pedestrian AP/APH 3D} \\
& L2 & L1 & L2 & L1 & L2 \\
\midrule
P.Pillars~\cite{lang2019pointpillars} \dag  &55.1 & 68.6/68.1 &  60.5/60.1 &  68.0/55.5 & 61.4/50.1 \\
CenterPoint~\cite{yin2021centerpoint}  & 69.1 &80.2/79.7 & 72.2/71.8 & 78.3/72.1 & 72.2/66.4 \\
RSN\_3f~\cite{sun2021rsn} &69.7 & 80.7/80.3 & 71.9/71.6 & 78.9/75.6 & 70.7/67.8 \\
PVRCNN++~\cite{shi2021pv++} & 71.2  &81.6/81.2  & 73.9/73.5 & 80.4/75.0 & 74.1/69.0 \\
SST\_TS\_3f~\cite{singlestride21}  &72.9&81.0/80.6 & 73.1/72.7 & 83.1/79.4 & 76.7/73.1 \\
SWFormer \cite{swformer}  & 73.4 & 82.9/82.5 & 75.0/74.7 & 82.1/78.1 & 75.9/72.1 \\
\textbf{SWFormer(+LA)}  &\textbf{74.8}& \textbf{84.0/83.6} & \textbf{76.3/76.0}& \textbf{83.1/79.3} & \textbf{77.2/73.5} \\
\bottomrule
\end{tabular}
}
\centering
\caption{\textbf{WOD test-set results.} \lidaraug{} (LA) significantly improves detection performance for SWFormer, and achieves new state-of-the-art mAPH L2.}
\label{tab:test_result}
\end{table}

\subsection{\lidaraug{} enables better model scaling}
Scaling up model capacity is a common approach to achieve better performance, but large 3D object detectors often suffer from overfitting. \autoref{tab:upillar-scale} shows scaling results, where UPillars-L is a larger model with more layers and channels than UPillars, detailed in \autoref{fig:UPillars}.

Here, we adopt the strong data augmentations used in the latest SWFormer as Baseline (see \autoref{sec:sota_swf}). As shown in \autoref{tab:upillar-scale}, with Baseline augmentations, UPillars-L does not benefit much from its significantly larger capacity. In fact, several metrics, such as Veh/Ped L1 AP, even become worse (e.g.  69.3/70.3  for UPillar-L vs 72.1/72.3 for UPillars). We  observe the training loss of UPillars-L is much smaller compared to loss of UPillars, indicating severe overfitting. 

On the other hand, \lidaraug{} achieves much better performance on larger models, especially on the most challenging metric, i.e., +7.3AP for 3D L2 mAPH as shown in \autoref{fig:scaling}.
Perhaps surprisingly, although the baseline UPillar (mAPH=57.8) is much worse than latest 3D detectors, the final performance of UPillar-L (+ \lidaraug{}) is actually competitive with the latest SWFormers, i.e., their mAPH are 71.0 vs. 72.8. This opens up new research opportunities on exploring much larger and higher performance 3D detectors in the future.

\begin{table}[!h]
\centering
\resizebox{0.48\textwidth}{!}{%
\begin{tabular}{c|cc|cc}
\toprule
\multirow{2}{*}{} & \multicolumn{2}{c|}{Veh/Ped AP L1} & \multicolumn{2}{c}{Veh/Ped APH L2} \\
& UPillars & UPillars-L & UPillars & UPillars-L \\
\midrule
BaseAugment & 72.1/72.3 & 69.3/70.3 & 63.5/52.1 & 61.0/59.0\\
\lidaraug{} & \textbf{77.1/77.5} & \textbf{79.5/81.6} & \textbf{68.5/58.9} & \textbf{71.5/70.5} \\
\bottomrule
\end{tabular}
}
\caption{\textbf{UPillars scaling results on WOD}.}
\label{tab:upillar-scale}
\vspace{-0.1in}
\end{table}

\subsection{\lidaraug{} supports different representations} \label{sec:multiview}
Different from 2D image models, 3D detectors are more diverse and could utilize different input representations due to the additional dimensionality and sparsity of point cloud data. Other than UPillars and SWFormer, which are both pillar-based architectures and taking 3D sparse points as inputs, we further demonstrate \lidaraug{} generalizes to other input representations. First, StarNet~\cite{ngiam2019starnet} is a point-based detector which directly processes raw points in 3D to detect objects.  RSN, on the other hand, utilize multi-view property of point clouds and takes both range images and 3D sparse points as inputs. However, due to the lack of multi-view data augmentations in prior works, RSN only utilize two simple augmentations, i.e. random flip and rotation.


\begin{table}[!h]
\centering
\resizebox{0.45\textwidth}{!}{%
\begin{tabular}{cc|cc}
\toprule
Model & Augmentation & Vehicle& Pedestrian \\
\midrule
\multirow{2}{*}{StarNet~\cite{ngiam2019starnet}} &  baseline & 58.2 & 71.9 \\
& \textbf{+\lidaraug{}} & \textbf{61.6} &\textbf{74.2}\\
\midrule
\multirow{2}{*}{RSN-1frame~\cite{sun2021rsn}} & baseline & 75.2 & 77.2 \\
& \textbf{+\lidaraug{}}  & \textbf{75.8}  &\textbf{79.0}\\
\midrule
\multirow{2}{*}{RSN-3frame~\cite{sun2021rsn}} & baseline & 77.0 & 79.1 \\
& \textbf{+\lidaraug{}}  &  \textbf{77.7}  &\textbf{80.6}\\
\midrule
\multirow{2}{*}{SWFormer~\cite{swformer}} & baseline & 79.4 & 82.9 \\
& \textbf{+\lidaraug{}}  &\textbf{80.9} & \textbf{84.4}\\
\bottomrule
\end{tabular}
}
\caption{\textbf{\lidaraug{} improves various models}. Startnet is a point-based detector. RSN is a range image and pillar-based detector. Results are WOD L1 AP.}
\label{tab:multiview}
\end{table}

\lidaraug{} is a general method which supports augmenting different views of point clouds, including range images, as explained in \autoref{subsec:range}. 
\autoref{tab:multiview} shows the performance of \lidaraug{} on point-based StarNet, range image based RSN, and transformer-based SWFormer. In general, our \lidaraug{} improves all kinds of 3D detectors, sometimes by a large margin.


\subsection{Abletion studies: comparing to other approaches. } \label{sec:compareppba}
In this section, we show \lidaraug{}  outperforms other common data augmentation approaches on UPillars. 

\textbf{Manually tuned data augmentation.}
Due to the complexity of search space scales exponentially with respect to the number of parameters, commonly used data augmentation strategies often consists of few data augmentation operations. Here, we benchmark two sets of data augmentation strategies used in training high-performance 3D detectors. First, we adopt random flip (probability 0.5) and rotation (probability 0.5, yaw angle uniformly sampled from [$-\pi/4$,  $\pi/4$]) data augmentations used in training RSN \cite{sun2021rsn}. Then we benchmark more advanced and stronger data augmentation strategy used in training SWFormer \cite{swformer}, detailed in \autoref{sec:sota_swf}. Our results show both data augmentation strategies significantly improved UPillars performances, about +10 AP for Vehicle and Pedestrian 3D L1 AP, when compared to the no augmentation baseline, shown in \autoref{tab:UPillars-LA}. However, tuning the data augmentation hyperparameters is challenging, e.g., if we only search 4 values for each hyperparameter, the number of searches of 5 hyperparameters exceeds 1000.

\begin{table}[h]
\centering
\resizebox{0.49\textwidth}{!}{%
\begin{tabular}{c|c|ll}
\toprule
UPillars (AP Level 1) &Hparams& Vehicle & Pedestrian  \\
\midrule
No Augmentation & - & 58.0 & 62.4 \\
Rotate \& Flip \cite{sun2021rsn}& 3 &70.8 & 70.0 \\
Rotate, Flip, Scale, Drop points \cite{swformer}& 5 &\textbf{72.1} & 72.3 \\
PPBA \cite{cheng2020improving}& 29 &71.6 & \textbf{72.6} \\
\midrule
\textbf{\lidaraug{}} & \textbf{2} &\textbf{77.1} \small \textcolor{blue}{(+5.0)} &\textbf{77.5 }\small \textcolor{blue}{(+4.9)}\\
\bottomrule
\end{tabular}
}
\caption{\textbf{\lidaraug{} outperforms common data augmentation strategies.} UPillars L1 APs on Waymo Open Dataset \textit{validation set} are reported. \lidaraug{} requires the least number of hyperparameters (Hparams) but achieves the best results compared to manually designed and automl-based data augmentations strategies.}
\label{tab:UPillars-LA}
\end{table}

\textbf{AutoML-based data augmentation.}
To alleviate the challenge of exponentially large search space, population-based training is proposed to tune hyperparameters in data augmentations online \cite{jaderberg2017population,ho2019population,cheng2020improving}. We follow the implementation of progressive-population based data augmentation (PPBA) \cite{cheng2020improving} and use the same sets of data augmentation policies and search space. We set population size 16, generation step 4000, perturbation and exploration rate to 0.2. Our results, in \autoref{tab:UPillars-LA}, show PPBA significant outperforms the no augmentation baseline. Despite PPBA introduces significantly more data augmentation policies, it is on par with manually tuned data augmentations, which only contains 4 policies.

We find the search space of PPBA is suboptimal after inspecting the search domain of each hyperparameter. For example, the maximum rotation angle for global rotation in the PPBA search space is $\pi/4$, a common value used for KITTI dataset. However, $\pi/4$ is insufficient compared to the tailored max rotation angle $\pi$ used in our \lidaraug{}.  Surprisingly, a single well-tuned global rotation augmentation achieves L1 mAP 73.3, shown in \autoref{tab:searchspace}, which outperforms PPBA with L1 mAP 72.1 over vehicle and pedestrian tasks.  Although PPBA algorithm is more efficient than grid search and contains diverse augmentation policies, the suboptimal search domain of rotation angle restricts the performance of PPBA, which highlights the importance of tailoring 3D detection search space.

\textbf{\lidaraug{}}
Alternatively, \lidaraug{} mitigates both the curse of dimensionality and suboptimal search domain issues by aligning and scaling the magnitude and probability of each data augmentation policy. This significantly reduces the search complexity (only 2 hyperparameters) while allowing exploration of a larger hyperparameter space. As indicate in \autoref{tab:UPillars-LA}, \lidaraug{} significantly outperforms both manually designed and AutoML-based data augmentation strategies by about 5 AP for both vehicle and pedestrian detection tasks and only requires a simple grid search of two hyperparameters.

\subsection{Generalize to nuScenes dataset} \label{sec:nuScenes}
To further validate our method, we evaluate \lidaraug{} on a different dataset: nuScenes~\cite{caesar2020nuscenes}. For simplicity, we adopt the same training settings as Waymo Open Dataset, but reduce the voxel size to 0.25 and the total training steps by half for faster training. We use the same baseline augmentation as SWFormer, and redefine \lidaraug{} search space for nuScenes following \autoref{sec:define_searchspace}. \autoref{tab:nuscenes} shows \lidaraug{} is a general approach, which outperforms the baseline augmentation by a large margin on nuScenes.

\begin{table}[!h]
\centering
\resizebox{0.45\textwidth}{!}{%
\begin{tabular}{c|ll}
\toprule
UPillars  & mAP & NDS  \\
\midrule
Rotate, Flip, Scale, Drop points \cite{swformer} & 40.6 & 48.2\\
\textbf{\lidaraug{}}  &\textbf{46.7} \small \textcolor{blue}{(+6.1)} &\textbf{53.4} \small \textcolor{blue}{(+5.2)}\\
\bottomrule
\end{tabular}
}
\caption{\textbf{nuScenes validation-set results}.}
\label{tab:nuscenes}
\vspace{-0.1in}
\end{table}

\section{Conclusion}
In this paper, we propose \emph{\lidaraug{}}, a scalable and effective 3D augmentation approach for 3D object detection. Based on the insight that 3D data augmentations are sensitive to model architecture and capacity, we propose a simplified search space, which contains two hyperparameters to control a diverse set of augmentations. \lidaraug{} outperforms both manually tuned and existing search-based data augmentation strategies by a large margin. Extensive studies show that \lidaraug{} generalizes to convolution and attention-based architectures, as well as point-based and range-based input representations. More importantly, \lidaraug{} significantly simplifies the search process for 3D data augmentations and opens up exciting new research opportunities, such as model scaling in 3D detection. With \lidaraug{}, we demonstrate new state-of-the-art 3D  detection results on the challenging Waymo Open Dataset.









\bibliography{root}
\bibliographystyle{IEEEtran}

\end{document}